\documentclass[11pt]{article}

\usepackage[preprint]{acl}

\usepackage{times}
\usepackage{latexsym}

\usepackage[T1]{fontenc}

\usepackage[utf8]{inputenc}

\usepackage{silence}
\usepackage{microtype}

\usepackage{inconsolata}

\usepackage{hyperref}       
\usepackage{url}            
\usepackage{booktabs}       
\usepackage{amsfonts}       
\usepackage{nicefrac}       
\usepackage{xcolor}         
\usepackage[plainruled,vlined,linesnumbered]{algorithm2e}
\usepackage{amsmath}
\usepackage{graphicx}
\usepackage{array}
\usepackage{amssymb}
\usepackage[capitalize, noabbrev]{cleveref} 
\usepackage{color}
\usepackage{tabularray}
\usepackage[table]{xcolor}
\usepackage{multirow}
\usepackage{listings}
\usepackage{subcaption}
\usepackage{pifont}
\usepackage{colortbl}
\usepackage{bm}
\usepackage{mathtools} 
\usepackage{comment}
\usepackage[normalem]{ulem}

\DeclarePairedDelimiter{\norm}{\lVert}{\rVert}

\crefname{lstlisting}{Listing}{Listings}

\newcommand{\ie}{\textit{i}.\textit{e}., }

\usepackage{makecell}

\newcommand{\ourrow}{\rowcolor{gray!15}}
\newcommand{\cc}{\cellcolor{gray!15}}

\title{ 
Layer-wise Curriculum Learning for Efficient LLM Compression
} 

\author{
Donggeon Lee%
~\;~\;~Dooyeon Na%
~\;~\;~Seungmin Oh%
~\;~\;~Jongbin Ryu\thanks{Corresponding author.}\\
{Ajou University, South Korea}\\
\texttt{\{\href{mailto:donggeon\_lee@ajou.ac.kr}{donggeon\_lee},
\href{mailto:dooyeonna@ajou.ac.kr}{dooyeonna},
\href{mailto:seungminoh@ajou.ac.kr}{seungminoh},
\href{mailto:jongbinryu@ajou.ac.kr}{jongbinryu}\}@ajou.ac.kr}}

\begin{document}
\maketitle

\begin{abstract}
In this paper, we introduce layer-wise curriculum learning for efficient LLM compression.
The proposed method facilitates the knowledge transfer from the teacher model to the student model, utilizing a curriculum learning approach that begins with easier optimization tasks and progressively tackles harder ones.
In order to adopt the layer-wise learning in LLM compression, we partition the whole model into multiple segments consisting of layers, thereby enabling more computationally efficient knowledge transfer for LLMs.
Based on our theoretical analysis of cumulative error phenomenon, layer-wise curriculum learning accelerates convergence while stabilizing the knowledge transfer process.
In addition, we present a feature caching method with a multi-threading strategy to efficiently address feature misalignment across layers, maximizing GPU utilization.
Consequently, our method exhibits advanced model compression performance, as well as high computational efficiency in terms of minimized memory usage and short training hours.
Experiments on multiple datasets show that the proposed method achieves state-of-the-art performance while reducing GPU memory usage and training hours by more than 50\% on BERT and GPT-2.
Moreover, it outperforms the other pruning methods on LLaMA-family and Qwen models under the same training hours, with a lower GPU memory footprint.
Official code is available at \url{https://github.com/mmai-laboratory/layer_wise_curriculum}.
\end{abstract}

\section{Introduction}
\label{introduction}
As language models grow in parameter size, they demonstrate significantly enhanced performance for various tasks~\cite{radford2019language,devlin2019bert,touvron2023llama,grattafiori2024llama}. 
However, deploying increasingly large models requires substantial hardware resources, creating serious computational challenges.
Consequently, recent work has focused on compressing language models to reduce parameter counts of layers while preserving performance~\cite{sanh2019distilbert,wang2020minilm,dasguptaimproving,liang2023less,chen2024streamlining,yang2024laco}.
There are two main approaches for this layer reduction: 
1) knowledge distillation (KD), which trains a student model with knowledge transfer from a teacher model, and 
2) pruning, which removes less important layer parameters without the knowledge transfer.
Although both aim to compress models while preserving performance, they differ in how compression is performed.
KD benefits from explicit knowledge transfer with massive resource overhead, while pruning is more efficient but does not exploit teacher guidance during compression.

\begin{figure}[t!]
    \subfloat[Cosine distance]{\includegraphics[width=0.5\columnwidth]{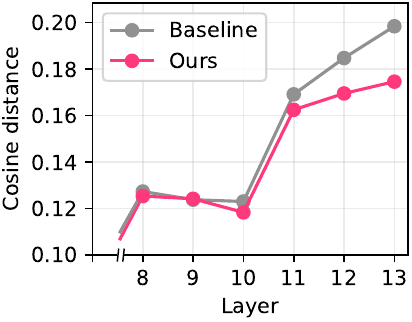}}
    \subfloat[Performance]{\includegraphics[width=0.5\columnwidth]{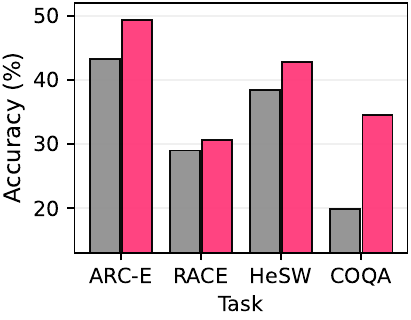}}
    \caption{
    Empirical evaluation of layer-wise cosine distance and downstream performance.
    (a) cosine distance illustrates the evolution of error relative to the teacher model in a layer-wise manner,
    while (b) performance reports task accuracy, which quantifies the quality of final representation.
    Baseline denotes the conventional layer-wise knowledge transfer method, which employs uniform optimization across all layers.
    }
    \label{fig:cosine-distance-and-performance}
\end{figure}

To achieve stable compression performance under tight memory budgets without sacrificing training speed, we build on a layer-wise knowledge transfer approach that inherits the strengths of both knowledge distillation (KD) and pruning.
Our approach transfers knowledge in a partitioned, layer-wise manner, significantly reducing training cost while retaining strong compression quality.
As illustrated in \cref{fig:cosine-distance-and-performance}, we empirically observe an error accumulation phenomenon as layer depth increases, which stems from feature misalignment between student layers. 
This misalignment induces a performance degradation in final representation, ultimately limiting the stability of compressed models.
We formalize this depth-wise error accumulation via Lipschitz continuity, explaining why deeper layers become increasingly error-prone and why uniform layer-wise optimization is suboptimal.
Guided by this analysis, we introduce layer-wise curriculum learning, which adopts a curriculum that progressively shifts the optimization focus from all layers to deeper layers.
This design directly targets the root cause of the cumulative error and yields a more effective use of a fixed optimization budget.
In addition, to resolve the feature misalignment across layers efficiently, we introduce a feature caching method.
This feature caching method suppresses unnecessary computational overhead by implementing multi-threaded knowledge transfer with parallel processing.
Furthermore, we employ a direction-based loss that leverages directionality of feature vectors to facilitate more robust and efficient knowledge transfer compared to conventional magnitude-based losses.
Our framework makes the following contributions:
\begin{itemize}
    \item We theoretically interpret the cumulative error phenomenon by formalizing it under Lipschitz continuity, explaining why feature misalignment becomes more severe in deeper layers.
    \item We propose layer-wise curriculum learning that progressively reallocates more optimization budget toward deeper, more error-prone layers as training proceeds.
    \item We introduce an efficient pipeline via feature caching and multi-threading strategy, removing redundancy and accelerating convergence.
\end{itemize}
To validate the robustness and efficiency of the proposed approach, we perform comparative experiments with state-of-the-art methods across four language models, such as BERT~\cite{devlin2019bert}, GPT-2~\cite{radford2019language}, LLaMA-family~\cite{touvron2023llama,grattafiori2024llama} and Qwen~\cite{yang2024qwen25}.
The results indicate that our approach performs well while converging approximately 1.7 times faster.
We also demonstrate that our proposed framework runs effectively on memory-constrained environments.

\begin{figure*}
  \centering
  \includegraphics[width=0.95\textwidth]{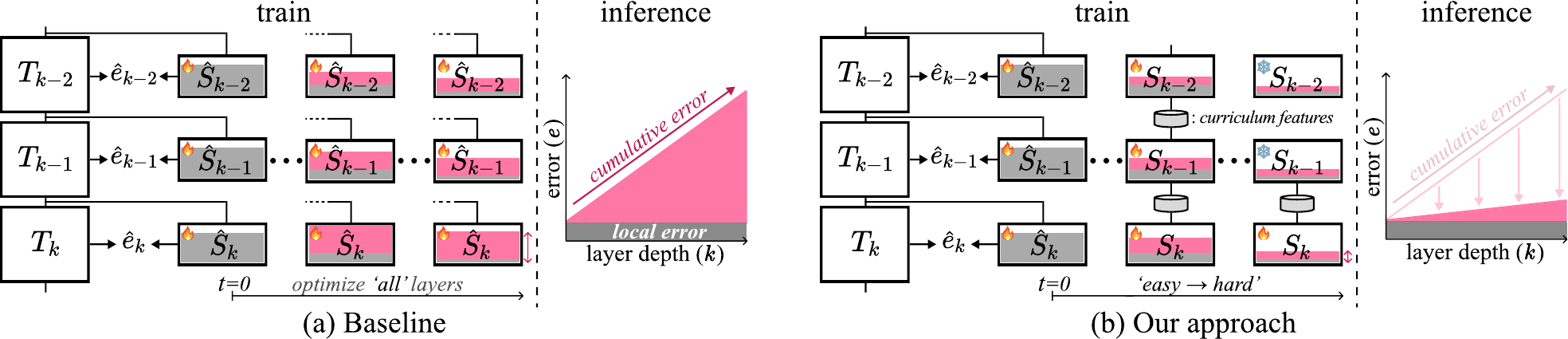}
  \caption{
  Workflow of the proposed method.
  While (a) baseline employs uniform learning under the same total training hours as ours, (b) our approach progressively allocates more optimization budgets toward deeper, error-prone layers, thereby effectively mitigating the cumulative error that builds up with depth of the layers.
  }
  \label{fig:workflow}
\end{figure*}

\section{Related Work}
\label{sec:related_work}
\subsection{Knowledge Distillation (KD)}
Knowledge distillation~\cite{hinton2015distilling} facilitates model compression by guiding smaller student networks to replicate the functionality of larger teacher networks.
DistilBERT, DistilGPT~\cite{sanh2019distilbert}, and MiniLM~\cite{wang2020minilm} train student networks using cross-entropy (CE) loss, Kullback-Leibler (KL) divergence, and cosine distance. 
CKA~\cite{dasguptaimproving} aligns student and teacher representations using centered kernel alignment.
TED~\cite{liang2023less} introduces task-aware filters at each layer to train the student networks in mimicking the features of the teacher.
Although KD often achieves strong compression performance, it demands additional memory overhead because both teacher and student networks must be fully loaded on each GPU.

To mitigate this limitation, layer-wise KD partitions models into segments and performs localized distillation, thereby reducing memory usage.
For example, \citet{blakeney2020parallel, jang2023pipe} propose optimized pipelines for decoupled layer-wise distillation.
\citet{chang2022distilhubert, kim2024ladimo} apply the layer-wise distillation to the speech representation and mixture-of-expert model.
\citet{yu2024decoupling} links teacher and student layers via connectors, but performs distillation by comparing the final logits of the entire network.
These methods enable more fine-grained knowledge transfer by supervising individual layers of the student model.
Building on these benefits, we introduce a new knowledge transfer method that improves with curriculum learning and feature caching methods.

\subsection{Pruning}
Pruning compresses models by identifying and removing redundant parameters or layers from pre-trained networks. 
LLM-Pruner~\cite{ma2023llm} removes unimportant structures based on gradient information, while LaCo~\cite{yang2024laco} collapses groups of layers by calculating the similarity of the final output.
Similarly, Streamline~\cite{chen2024streamlining} removes a redundant layer based on output similarity and replaces it with a trained lightweight network.
Although pruning is more efficient than KD, its performance degrades sharply as the compression ratio increases.
In contrast, our approach maintains stable performance even at high compression rates by incorporating teacher-guided knowledge transfer during compression.

\section{Method}
\label{sec:methodology}
This section presents the proposed framework, organized as follows.
We first analyze the cumulative error phenomenon inherent in layer-wise knowledge transfer based on Lipschitz continuity.
Subsequently, we propose a layer-wise curriculum learning method aimed at reducing the cumulative error, along with a feature caching method with a multi-threading strategy to effectively address the feature misalignment across layers.
Building upon these foundations, we introduce a unified compression framework that facilitates robust knowledge transfer through a layer-wise curriculum learning approach, integrated with an efficient feature caching method.
PyTorch-like pseudocode of the proposed framework is provided in \cref{sec:algorithm-of-entire-procedure}.
The detailed workflow of layer-wise curriculum learning is illustrated in \cref{fig:workflow}, while the end-to-end pipeline of our framework is detailed in \cref{fig:caching_multi-thread}.

\subsection{Knowledge Transfer}
\subsubsection{Cumulative Error}
\label{sec:cumulative-error}
Our approach is grounded in the cumulative error associated with layer-wise knowledge transfer.
As shown in \cref{fig:workflow}, learning each layer independently leads to feature misalignment across student layers.
To address this limitation, it is essential to analyze the error of the learning approach.
We first establish an ideal local optimization error between the features of teacher and student layers as:
\begin{equation}
\label{eq:eval_error}
    e_k = \norm{T_{k} - S_{k}},
\end{equation}
where $T_{k}$ and $S_{k}$ denote the output features of $k_{th}$ partitioned layers in the teacher and student models, respectively.
It captures the actual representation gap manifested at inference time, when student layers are composed sequentially without teacher intervention.
In practice, however, this ideal objective is seldom attained via layer-wise knowledge transfer.
Instead, each student layer is conditioned on the feature of the corresponding teacher layer.
Consequently, the exact local optimization error of the layer-wise learning approach is defined as:
\begin{equation}
\label{eq:train_error}
    \hat{e}_k = \norm{T_{k} - \hat{S}_{k}},
\end{equation}
where $\hat{S}_k$ represents the output feature of $k_{th}$ student layer whose input feature is based on the teacher model.
This mismatch is the source of feature misalignment: each student layer is trained on teacher features but receives student features at inference time.
To analyze the relationship between $e_k$ and $\hat{e}_k$, we formulate an upper bound of error $e_k$ using the triangle inequality as:
\begin{equation}
\label{eq:eval_error_minkowski}
\begin{aligned}
    e_k \; \le \; \norm{\hat{S}_{k}\!\! - S_{k}} + \norm{T_{k}\!\! - \hat{S}_{k}}.
\end{aligned}
\end{equation}
To bound the first term, we assume that each student layer locally adheres to Lipschitz continuity, such that for all $u$ and $v$, $\norm{f(u)-f(v)} \le L \norm{u - v}$ is satisfied, where $L$ denotes the Lipschitz constant.
Since we utilize the output features following the normalization layer in our knowledge transfer process, it can be asserted that the Lipschitz continuity is satisfied locally, as demonstrated in previous studies~\cite{santurkar2018does, xiong2020layer, 10.5555/3692070.3692295}.
While the normalization process inherently constrains the feature representations to a bounded set, such a bounding effect stabilizes learning dynamics by mitigating the risk of gradient explosion and limiting the variance of the output scale.
To validate the local Lipschitz assumption of our practical learning regime, we provide an empirical analysis of the local Lipschitz constant and local sensitivity in \cref{sec:empirical-analysis-lipschitz}.
By defining $f_k(\cdot)$ as the function of $k_{th}$ student layer, the features under comparison can be expressed as $\hat{S}_k \! = \! f_k(T_{k-1})$ and $S_k \! = \! f_k(S_{k-1})$.
Consequently, based on the Lipschitz continuity, the upper bound of the first term in \cref{eq:eval_error_minkowski} is determined as:
\begin{equation}
\label{eq:mismatch_error_minkowski}
\begin{aligned}
    \norm{\hat{S}_{k} \! - \! S_{k}} \! \le \! L_k \norm{T_{k-1} \!\! - \! S_{k-1}},
\end{aligned}
\end{equation}
where $L_k$ denotes the Lipschitz constant in $k_{th}$ student layer.
Thus, we simplify \cref{eq:eval_error_minkowski} as:
\begin{equation}
\label{eq:upper_bound_error}
    e_k \; \le \; L_k e_{k-1} + \hat{e}_k,
\end{equation}
where we define $e_{k-1} \! = \! \norm{T_{k-1} \! - S_{k-1}}$ as well as $\hat{e}_k \! = \! \norm{T_k \! - \hat{S}_k}$.
\cref{eq:upper_bound_error} recursively aggregates the propagated errors from previous layers. 
Therefore, by unrolling the recursion starting from the first layer, we derive the cumulative error as:
\begin{equation}
\label{eq:upper_bound_error_unroll}
    e_k \; \le \; \prod_{i=2}^{k}L_{i}e_{1} + \sum_{j=2}^{k} \prod_{i=j+1}^{k}L_{i}\hat{e}_{j}.
\end{equation}
It demonstrates that the error grows with layer depth $k$, motivating an optimal learning strategy that takes into account the layer depth of a model.

\subsubsection{Layer-wise Curriculum Learning}
\label{sec:curriculum_learning}
As indicated by the error analysis in \cref{eq:upper_bound_error_unroll}, optimizing deeper layers (\ie larger $k$) presents a more challenging problem, because they are subject to a larger upper bound on the error.
To address this issue, we introduce layer-wise curriculum learning guided by a depth-aware optimization schedule.
This schedule gradually reallocates the optimization budget from all layers to deeper, more error-prone layers as training progresses.
We implement the curriculum learning schedule via a time-dependent loss function.
For $(t \! > \! 0)$, the curriculum loss is defined as:
\begin{equation}
\label{eq:loss_curriculum}
    \mathcal{L}(t) = \sum_{k=1}^{K} \lambda_k(t) e_{k}, \; \text{where} \; \lambda_k(t) \! \in \! \{0,1\},
\end{equation}
where $\lambda_k(t)$ denotes a binary indicator function that dynamically determines whether $k_{th}$ layer is active in the optimization at timestep $t$.
This indicator is controlled by $\psi(t)$, which is defined as the minimum depth of the active layers at timestep $t$:
\begin{equation}
    \lambda_k(t) = \mathbb{I}[k \ge \psi(t)].
\end{equation}
Accordingly, increasing $\psi(t)$ progressively removes shallow layers from optimization and shifts the remaining training budget toward deeper, harder ones.
At initialization $(t = 0)$, each student layer is optimized using the corresponding teacher feature:
\begin{equation}
    \mathcal{L}(0) = \sum_{k=1}^{K} \hat{e}_{k} \;\; \text{where} \; \forall k : \lambda_{k}(0) = 1.
\end{equation}
As training proceeds $(t > \! 0)$, $\psi(t)$ is increased monotonically so that the converged shallowest active layer is frozen and excluded from the optimization process, adhering to the curriculum learning depicted in \cref{eq:loss_curriculum}.
The training ultimately culminates in the final stage, where the optimization focus is solely concentrated on the single deepest layer.
Under this schedule, the cumulative error of each active layer $k \ge \psi(t)$ is bounded by:
\begin{equation}
\label{eq:upper_bound_layers}
    {e}_{k} \le \! \prod_{i=\psi(t)}^k \!\! L_{i} {e}_{\psi(t)-1} + \! \sum_{j=\psi(t)}^k \prod_{i=j+1}^k \!\! L_i \hat{e}_j.
\end{equation}
Compared with the conventional bound in \cref{eq:upper_bound_error_unroll}, this expression reflects our strategy of freezing shallow layers $k < \psi(t)$ once they have converged, thereby avoiding redundant optimization on already-stabilized subproblems.
This design fundamentally redefines how optimization resources should be allocated across depth.
Since shallow layers accumulate less propagated error than deeper ones, they typically converge more rapidly.
Once the shallow layers have sufficiently converged, further optimizing them may still reduce the final error, but the resulting gain is relatively limited.
By allocating more optimization budget to deeper layers, where more cumulative error remains, our method avoids wasting the training budget on subproblems that have already reached stability.
Under a fixed optimization budget, this depth-aware prioritization, which focuses on mitigating cumulative error, plays a decisive role in enhancing the final representation quality compared to the conventional methods.
Consequently, we improve model compression performance by leveraging a well-designed curriculum learning schedule, which effectively optimizes cumulative error while maintaining efficient resource utilization.
Furthermore, to mitigate the feature misalignment across student layers arising from independent knowledge transfer, we incorporate the efficient feature caching method, which improves layer-wise feature alignment with minimal computational redundancy.
A more detailed description of $\psi(t)$ for the schedule is provided in \cref{sec:psi-description}, while a detailed exposition of feature caching with multi-threading strategy is described in the following subsection.

\begin{figure}[t!]
  \centering
  \includegraphics[width=.95\columnwidth]{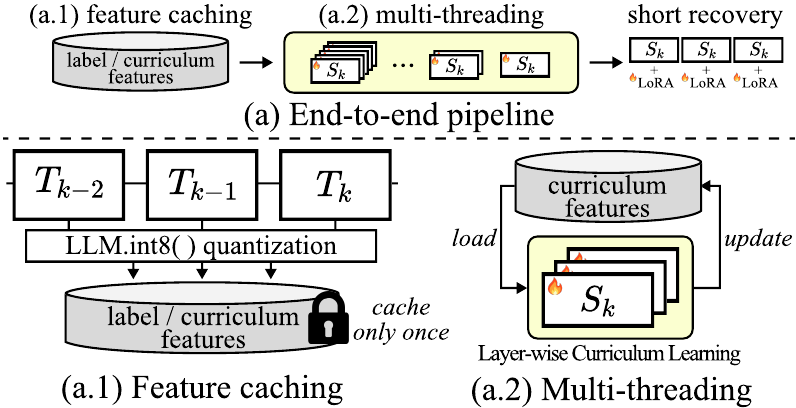}
  \caption{
  (a) end-to-end pipeline of the proposed method including feature caching with multi-threading strategy.
  (a.1) feature caching eliminates redundant teacher computations, while (a.2) multi-threading ensures efficient feature alignment between student layers.
  }
  \label{fig:caching_multi-thread}
\end{figure}

\subsection{Feature Caching with Multi-Threading}
We introduce feature caching with multi-threading strategy to accelerate layer-wise curriculum learning.
There are widely used multi-GPU acceleration tools to train LLMs, such as Distributed Data Parallel (DDP)~\cite{li2020pytorch} and Fully Sharded Data Parallel (FSDP)~\cite{zhao2023pytorch}.
DDP improves computational throughput but is memory-inefficient because it replicates full model on every device, while FSDP reduces memory usage via sharding at the cost of communication overhead.
However, multi-GPU execution is not an essential requirement of our LLM learning approach.

Our strategy addresses these limitations by caching forward-propagated features and partitioning layers across individual GPUs, thereby significantly reducing redundant computation and memory footprint.
As shown in \cref{fig:caching_multi-thread}, the features derived from the teacher model, which serve as labels for optimization, are cached only once at the beginning of the training process and subsequently reused.
The curriculum features, which feed into the student layers, are sourced from the teacher model only at the initial step $(t \! = \! 0)$, and dynamically updated from the student layers for all subsequent steps $(t \! > \! 0)$.
Beyond resource efficiency, this design is also theoretically grounded in error optimization.
As formalized in \cref{eq:upper_bound_error}, optimizing the error in $(k\!-\!1)_{th}$ layer lowers the upper bound of the error at the subsequent $k_{th}$ layer.
This hierarchical dependency allows us to optimize layer-wise error in parallel at each step, facilitating both rapid convergence and systematic management of the error propagation.
Furthermore, to improve storage efficiency, we apply LLM.int8()~\cite{10.5555/3600270.3602468} quantization to the cached features of LLaMA-family~\cite{touvron2023llama,grattafiori2024llama} and Qwen~\cite{yang2024qwen25}.
It strategically quantizes only non-outlier feature values, thereby minimizing information loss while preserving performance during knowledge transfer process.
Consequently, layer-wise curriculum learning leverages the feature caching with multi-threading strategy to efficiently align the features across student layers while significantly reducing overall training resource usage.
We provide detailed comparisons regarding training resource usage against the conventional methods in \cref{sec:gpu-utilization}.

\begin{table*}[t!]
  \centering
  \setlength{\tabcolsep}{4pt}
  \scriptsize
  \begin{tabular}{l|c|c|ccccccc|c}
    \toprule
                 & Train-hour (H) & Mem. (\%)  & COLA          & SST-2         & MRPC          & RTE           & STSB          & MNLI          & QNLI          & AVG   \\
    \midrule
    BERT$_{768 \times 12}$       & > 500        &  100.0   & 52.1          & 93.5          & 88.9          & 66.4          & 87.1          & 84.6          & 90.5          & 80.4 \\
    \midrule
    DistilBERT$_{768 \times 6}$  & 360   &  111.4   & 49.0          & \textbf{92.3}          & 86.9          & 58.4          & 81.3          & 82.6          & 88.8          & 77.0 \\
    MiniLM$_{768 \times 6}$      & 360   &  111.4   & 49.2          & 92.0          & 88.4          & 65.1          & 85.0          & 83.0          & \textbf{90.1} & 79.0 \\
    CKABERT$_{512 \times 12}$    & 200   &  105.0   & 50.2          & \textbf{92.3} & 87.8          & 63.0          & 84.9          & \textbf{88.5} & 90.0          & 79.5 \\
    \ourrow Ours$_{768 \times 6}$  & 160          &  45.4    & \textbf{52.3} & 92.2          & \textbf{89.3} & \textbf{66.8} & \textbf{87.4} & 82.2          & \textbf{90.1} & \textbf{80.0}  \\
    \bottomrule
  \end{tabular}
\caption{Experimental results of the KD methods on BERT. Train-hour is measured by \#GPU $\times$ Hour and Mem. denotes usage of the GPU memory during the compression process.}
\label{tab:bert-performance}
\end{table*}

\begin{table*}[t!]
  \centering
  \scriptsize
  \setlength{\tabcolsep}{8pt}
  \begin{tabular}{l|c|c|cccc}
    \toprule
    & \multirow{2}{*}{Train-hour (H)} & \multirow{2}{*}{Mem. (\%)} & OpenWebText & WikiText-103 & \multicolumn{2}{c}{LAMBADA} \\
    &  &  & ppl $\downarrow$ & ppl $\downarrow$ & ppl $\downarrow$ & acc $\uparrow$ \\
    \midrule
    GPT-2$_{768 \times 12}$             &  > 1000      &  100.0    & 23.1             & 37.5            & 35.1         & 46.0   \\
    \midrule
    DistilGPT-2$_{768 \times 6}$        &  400  &  121.9    & 29.1             & 49.0            & 87.9         & 22.9   \\
    LWD$_{768 \times 6}$                &  400  &  121.9    & 29.7             & 51.9            & 91.9         & 22.0   \\
    TED$_{768 \times 6}$                &  450  &  > 121.9  & 28.5             & 48.1            & 87.2         & 23.0   \\
    \ourrow Ours$_{768 \times 6}$       &  160  &  32.9     & \textbf{26.4}    & \textbf{46.3}   & \textbf{69.4}& \textbf{27.9}   \\
    \bottomrule
  \end{tabular}     
\caption{
Experimental results of the KD methods on GPT-2.
The ppl and acc denote perplexity and accuracy.
}
\label{tab:gpt2-performance}
\end{table*}

\subsection{Direction-based Loss}
\label{sec:direction-alignment-loss}
To effectively facilitate semantic knowledge transfer between the features of the teacher and student models, we categorize the objectives into magnitude-based and direction-based losses.
The magnitude-based losses minimize absolute magnitude discrepancies between features, but they inherently emphasize scale differences and often overlook the directionality of the feature vectors.
In contrast, direction-based losses focus on the directional relationship of features, which more directly reflects representational similarity.
Given that modern architectures extensively utilize normalization layers to stabilize feature scales, focusing on directionality serves a more robust and scale-invariant knowledge transfer for capturing the semantic relationship.
Accordingly, we employ a cosine similarity loss function to prioritize directional semantics, encouraging each student layer to capture the angular representations of its corresponding teacher layer.
This design not only accelerates convergence during the short training hours, but also preserves the core representations of knowledge being transferred.
We validate our approach in \cref{sec:ablation-loss-function}.

\section{Experiments}
\label{sec:experiments}

\subsection{Experimental Settings}
\textbf{Knowledge Distillation.} 
We evaluate KD on two foundational models: BERT~\cite{devlin2019bert} and GPT-2~\cite{radford2019language}.
For BERT, we compare our method against DistilBERT~\cite{sanh2019distilbert}, MiniLM~\cite{wang2020minilm}, and CKABERT~\cite{dasguptaimproving}.
For GPT-2, we benchmark our approach against Layer-Wise Distillation (LWD), DistilGPT~\cite{sanh2019distilbert}, and TED~\cite{liang2023less}.
In all KD experiments except CKABERT, the number of layers is reduced from 12 to 6.
In CKABERT, the hidden dimension is reduced from 768 to 512, while the number of total layers remains unchanged.

\noindent \textbf{Pruning.}
We compare the proposed method against three pruning methods: LLM-Pruner~\cite{ma2023llm}, LaCo~\cite{yang2024laco}, and Streamline~\cite{chen2024streamlining}.
We evaluate under pruning ratios of approximately 25\% and 50\% on LLaMA2-7B, LLaMA2-13B~\cite{touvron2023llama}, LLaMA3-3B, LLaMA3-8B~\cite{grattafiori2024llama}, Qwen2.5-7B, and Qwen2.5-14B~\cite{yang2024qwen25}.
Following the pruning process, a short recovery stage is conducted using LoRA~\cite{hu2022lora}.
All pruning experiments involve 24-48 GPU $\times$ Hour for knowledge transfer and 4-8 GPU $\times$ Hour for the recovery stage.
To ensure a fair comparison, we train all methods for the same training duration.
Additional implementation and evaluation details are provided in \cref{sec:appendix-experiment-setup}.

\subsection{Experimental Results}

\subsubsection{Knowledge Distillation}
\label{sec:knowledge-distillation-results}
\cref{tab:bert-performance,tab:gpt2-performance} present the experimental results of KD methods on BERT and GPT-2.
Conventional KD methods typically require both teacher and student models to be loaded onto the GPU, often resulting in significantly higher memory consumption than training from scratch.
In contrast, our method loads only the required layers to each GPU due to layer-wise learning.
Across both tables, our method reduces memory usage by more than 50\% compared with the other methods while achieving better performance.
Moreover, \cref{fig:compare-perplexity-trend} in Appendix shows that the proposed method converges 1.7 times faster than other conventional methods.

\begin{table*}[t!]
  \centering
  \setlength{\tabcolsep}{3pt}
  \scriptsize
  \begin{tabular}{c|l|c|c|cccccccccc|c}
    \toprule
                        & Method & Ratio (\%) & Mem. (\%) & ARC-E & ARC-C & RACE & BoolQ & WnGd & HeSW & PIQA & MathQA & COQA & MMLU & AVG\\
    \midrule
    \multirow{9}{*}{LLaMA2-7B} & Dense$_{32}$ & 0.0 & 100.0 & 74.5 & 46.2 & 39.6 & 77.7 & 69.1 & 76.0 & 79.1 & 28.2 & 76.8 & 41.2 & 60.8 \\
    & LLM-Pruner$_{32}$       & 27.2 & 50.2 & 50.5 & 34.5 & 33.1 & 65.7 & 57.9 & 64.1 & 72.1 & 22.2 & 26.8 & 25.4 & 45.2 \\
    & LaCo$_{23}$             & 27.2 & 74.9 & 60.2 & 34.8 & 37.2 & 67.7 & 61.7 & 65.5 & \textbf{74.6} & \textbf{24.9} & 63.1 & 23.1 & 51.3 \\
    & Streamline$_{23}$       & 27.2 & 28.3 & 59.2 & 36.0 & \textbf{38.5} & 67.5 & \textbf{67.4} & 63.2 & 69.8 & 24.3 & 65.2 & 29.9 & 52.1 \\
    & \cc {Ours$_{23}$}         & \cc {27.2} &\cc {21.2} & \cc {\textbf{61.9}} & \cc {\textbf{36.4}} & \cc {37.6} & \cc {\textbf{76.4}} & \cc {67.1} & \cc {\textbf{67.2}} & \cc {73.7} & \cc {24.7} & \cc {\textbf{71.9}} & \cc {\textbf{31.6}} & \cc {\textbf{54.9}} \\
    & LLM-Pruner$_{32}$       & 46.7 & 50.2 & 43.1 & 27.6 & 29.9 & 59.7 & \textbf{54.0} & 46.6 & 66.5 & 21.7 & 14.7 & 23.5 & 38.7 \\
    & LaCo$_{16}$             & 48.1 & 74.9 & 45.8 & 26.9 & 30.1 & 56.1 & 53.5 & 47.3 & 67.2 & \textbf{22.8} & 36.3 & 23.1 & 40.9 \\
    & Streamline$_{16}$       & 48.1 & 28.3 & 39.7 & 23.9 & 27.3 & 61.3 & 53.1 & 26.5 & 63.8 & 21.0 & 7.7 & 23.1 & 34.7 \\
    & \cc {Ours$_{16}$}     & \cc {48.1} & \cc {19.6} & \cc {\textbf{48.0}} & \cc {\textbf{28.6}} & \cc {\textbf{33.7}} & \cc {\textbf{61.6}} & \cc {53.2} & \cc {\textbf{51.1}} & \cc {\textbf{68.3}} & \cc {22.4} & \cc {\textbf{41.7}} & \cc {\textbf{25.5}} & \cc {\textbf{43.4}} \\
    \midrule
    \multirow{9}{*}{LLaMA2-13B} & Dense$_{40}$ & 0.0 & 100.0 & 77.5 & 49.1 & 40.5 & 80.6 & 72.4 & 79.4 & 80.5 & 31.8 & 78.3 & 50.5 & 64.1 \\
    & LLM-Pruner$_{40}$       & 24.2 & 51.6 & 67.9 & 43.9 & 36.7 & 68.9 & 66.3 & 72.7 & 77.3 & 24.3 & 39.0 & 39.3 & 53.6 \\
    & LaCo$_{30}$             & 24.4 & 73.9 & 69.8 & 41.2 & 37.9 & 69.7 & 63.0 & 69.5 & \textbf{77.8} & 26.5 & 63.3 & 43.0 & 56.2 \\
    & Streamline$_{30}$       & 24.4 & 29.0 & 67.2 & 43.2 & \textbf{40.0} & 62.8 & \textbf{70.6} & 71.1 & 74.4 & 26.3 & 73.1 & 48.4 & 57.7 \\
    & \cc  {Ours$_{30}$}     & \cc {24.4} & \cc {23.5} & \cc {\textbf{70.7}} & \cc {\textbf{44.3}} & \cc {39.1} & \cc {\textbf{72.1}} & \cc {68.4} & \cc {\textbf{74.0}} & \cc {77.1} & \cc {\textbf{27.8}} & \cc {\textbf{74.4}} & \cc {\textbf{50.2}} & \cc {\textbf{59.8}} \\
    & LLM-Pruner$_{40}$       & 48.5 & 51.6 & 47.2 & 32.9 & 30.7 & 61.4 & 53.7 & 52.9 & 64.6 & 21.5 & 22.1 & 24.8 & 41.2 \\
    & LaCo$_{20}$             & 48.8 & 73.9 & 52.1 & \textbf{34.3} & 28.5 & 62.7 & 62.4 & 57.4 & 70.1 & \textbf{23.7} & 53.5 & 29.8 & 47.5 \\
    & Streamline$_{20}$       & 48.8 & 29.0 & 46.8 & 30.9 & 35.4 & 62.5 & \textbf{65.1} & 50.7 & 65.1 & 23.3 & 51.3 & \textbf{47.1} & 47.8 \\
    & \cc {Ours$_{20}$}     & \cc {48.8} & \cc {20.9} & \cc {\textbf{55.8}} & \cc {32.3} & \cc {\textbf{37.4}} & \cc {\textbf{65.2}} & \cc {64.1} & \cc {\textbf{59.8}} & \cc {\textbf{70.4}} & \cc {22.8} & \cc {\textbf{66.6}} & \cc {30.1} & \cc {\textbf{50.5}} \\
    \midrule
    \midrule
    \multirow{9}{*}{LLaMA3-3B} & Dense$_{28}$        & 0.0 & 100.0 & 71.8 & 46.2 & 39.8 & 73.0 & 69.1 & 73.6 & 77.3 & 34.4 & 78.5 & 54.1 & 61.8\\
    & LLM-Pruner$_{28}$       & 28.0 & 53.9 & 48.3 & 32.1 & 30.9 & 57.0 & 53.3 & 52.4 & 66.4 & 25.2 & 34.3 & 25.7 & 42.5 \\
    & LaCo$_{19}$             & 28.0 & 90.5 & 52.6 & 31.6 & 34.6 & 63.0 & 59.4 & 54.4 & 67.4 & 24.9 & 57.8 & 24.9 & 47.0\\
    & Streamline$_{19}$       & 28.0 & 32.1 & 47.2 & 31.0 & 32.0 & \textbf{67.5} & \textbf{65.3} & 48.9 & 64.2 & 22.5 & 59.8 & \textbf{51.8} & 49.0\\
    & \cc {Ours$_{19}$}     & \cc {28.0} & \cc {28.9} & \cc {\textbf{58.4}} & \cc {\textbf{36.4}} & \cc {\textbf{36.4}} & \cc {64.7} & \cc {64.0} & \cc {\textbf{59.8}} & \cc {\textbf{71.7}} & \cc {\textbf{25.2}} & \cc {\textbf{66.7}} & \cc {31.8} & \cc {\textbf{51.5}} \\
    & LLM-Pruner$_{28}$       & 43.9 & 53.9 & 42.8 & 26.0 & 28.6 & 55.1 & 50.0 & 39.5 & 64.3 & 24.0 & 16.6 & \textbf{24.3} & 37.2\\
    & LaCo$_{14}$             & 43.9 & 90.5 & 45.5 & \textbf{26.4} & 29.2 & 57.4 & 51.2 & 41.8 & 64.8 & 24.4 & 25.7 & 23.2 & 39.0\\
    & Streamline$_{14}$       & 43.9 & 32.1 & 44.2 & 24.4 & 26.9 & \textbf{61.5} & 51.3 & 36.6 & 65.7 & 23.5 & 9.9 & 22.9 & 36.8\\
    & \cc {Ours$_{14}$}         & \cc {43.9} & \cc {28.1} & \cc {\textbf{49.0}} & \cc {25.8} & \cc {\textbf{30.0}} & \cc {59.6} & \cc {\textbf{53.4}} & \cc {\textbf{43.4}} & \cc {\textbf{66.5}} & \cc {\textbf{25.0}} & \cc {\textbf{30.9}} & \cc {23.0} & \cc {\textbf{40.7}} \\
    \midrule
    \multirow{9}{*}{LLaMA3-8B} & Dense$_{32}$        & 0.0 & 100.0 & 81.2 & 53.2 & 38.9 & 82.0 & 74.2 & 79.0 & 81.1 & 39.7 & 80.4 & 63.1 & 67.3 \\
    & LLM-Pruner$_{32}$       & 24.4 & 50.4 & 56.1 & 36.4 & 33.6 & 66.0 & 59.5 & 64.2 & 70.2 & 26.7 & 40.3 & 31.2 & 48.4 \\
    & LaCo$_{23}$             & 24.4 & 73.5 & 65.4 & 40.4 & 36.2 & 66.5 & 64.1 & 67.4 & 75.6 & 28.5 & 70.6 & 41.0 & 55.5 \\
    & Streamline$_{23}$       & 24.4 & 28.0 & 67.4 & 41.6 & \textbf{38.2} & 75.8 & 71.1 & 65.5 & 72.0 & 30.1 & 68.4 & \textbf{62.1} & 59.2 \\
    & \cc {Ours$_{23}$}         & \cc {24.4} & \cc {19.9} & \cc {\textbf{70.2}} & \cc {\textbf{43.7}} & \cc {37.9} & \cc {\textbf{81.3}} & \cc {\textbf{72.2}} & \cc {\textbf{70.2}} & \cc {\textbf{75.8}} & \cc {\textbf{31.2}} & \cc {\textbf{79.5}} & \cc {61.2} & \cc {\textbf{62.3}}  \\
    & LLM-Pruner$_{32}$       & 43.5 & 50.4 & 47.1 & 30.3 & 31.0 & 57.2 & 54.4 & 48.9 & 66.3 & 24.5 & 22.1 & 24.5 & 40.6 \\
    & LaCo$_{16}$             & 43.5 & 73.5 & 47.7 & 28.0 & 33.0 & 61.7 & 55.9 & 50.0 & 68.2 & 23.5 & 44.5 & 23.0 & 43.5 \\
    & Streamline$_{16}$       & 43.5 & 28.0 & 48.3 & 25.5 & 28.3 & 61.3 & 52.2 & 41.2 & 69.1 & 22.8 & 14.0 & 22.9 & 38.5 \\
    & \cc {Ours$_{16}$}         & \cc {43.5} & \cc {18.5} & \cc {\textbf{54.7}} & \cc {\textbf{30.5}} & \cc {\textbf{33.5}} & \cc {\textbf{61.9}} & \cc {\textbf{58.1}} & \cc {\textbf{52.7}} & \cc {\textbf{70.1}} & \cc {\textbf{23.8}} & \cc {\textbf{48.2}} & \cc {\textbf{24.6}} & \cc {\textbf{45.8}}   \\
    \bottomrule
  \end{tabular}
\caption{
Experimental results of pruning methods on the LLaMA-family. Each method uses the same training hours.
Training memory usage is reported without the feature caching mechanism to guarantee a fair comparison.
}
\label{tab:llama-performance-table}
\end{table*}
\begin{table*}[t!]
  \centering
  \setlength{\tabcolsep}{3.4pt}
  \scriptsize
  \begin{tabular}{c|l|c|c|cccccccccc|c}
    \toprule
                       & Method & Ratio (\%) & Mem. (\%) & ARC-E & ARC-C & RACE & BoolQ & WnGd & HeSW & PIQA & MathQA & COQA & MMLU & AVG\\
    \midrule
    \multirow{7}{*}{Qwen2.5-7B} & Dense$_{28}$ & 0.0 & 100.0 & 77.5 & 51.1 & 41.8 & 84.8 & 73.3 & 78.9 & 79.8 & 43.4 & 83.2 & 71.7 & 68.6 \\
    & LaCo$_{19}$             & 27.6 & 90.8 & 59.6 & 36.9 & 32.5 & 61.7 & 59.2 & 58.4 & 70.7 & \textbf{26.5} & 54.2 & \textbf{24.2} & 48.4 \\
    & Streamline$_{19}$       & 27.6 & 34.0 & 59.7 & 32.9 & 33.8 & 60.1 & 55.2 & 57.3 & \textbf{74.1} & \textbf{26.5} & 43.8 & 22.9 & 46.6 \\
    & \cc Ours$_{19}$         & \cc 27.6 &\cc 26.2 & \cc \textbf{60.8} & \cc \textbf{37.2} & \cc \textbf{35.9} & \cc \textbf{65.0} & \cc \textbf{63.2} & \cc \textbf{62.2} & \cc 72.7 & \cc 25.0 & \cc \textbf{64.1} & \cc 23.2 & \cc \textbf{50.9} \\
    & LaCo$_{14}$             & 42.9 & 90.8 & 38.6 & 24.4 & 28.1 & 61.7 & 50.0 & 35.0 & 60.9 & 21.6 & 26.9 & 23.0 & 37.0 \\
    & Streamline$_{14}$       & 42.9 & 34.0 & 46.9 & 26.1 & 29.2 & 61.2 & 53.1 & 43.1 & 67.2 & \textbf{23.6} & 17.2 & 22.9 & 39.1 \\
    & \cc Ours$_{14}$         & \cc 42.9 & \cc 24.2 & \cc \textbf{49.9} & \cc \textbf{26.6} & \cc \textbf{31.5} & \cc \textbf{62.6} & \cc \textbf{54.6} & \cc \textbf{47.1} & \cc \textbf{67.6} & \cc 23.3 & \cc \textbf{43.3} & \cc \textbf{23.2} & \cc \textbf{43.0}  \\
    \midrule
    \multirow{7}{*}{Qwen2.5-14B} & Dense$_{48}$ & 0.0 & 100.0 & 79.4 & 58.8 & 40.8 & 85.4 & 75.8 & 56.9 & 81.9 & 51.8 & 83.2 & 77.6 & 69.2 \\
    & LaCo$_{35}$             & 24.2 & 84.6 & 63.0 & 39.7 & 34.7 & \textbf{68.8} & 65.7 & 63.8 & 72.0 & 27.6 & 69.5 & 41.7 & 54.7 \\
    & Streamline$_{35}$       & 24.2 & 31.6 & 57.3 & 35.3 & 33.2 & 57.8 & 59.0 & 63.9 & \textbf{76.6} & 26.5 & 47.4 & 22.9 & 48.0 \\
    & \cc Ours$_{35}$         & \cc 24.2 &\cc 27.3 & \cc \textbf{71.8} & \cc \textbf{45.1} & \cc \textbf{37.3} & \cc 63.2 & \cc \textbf{71.2} & \cc \textbf{67.7} & \cc 74.4 & \cc \textbf{28.7} & \cc \textbf{77.6} & \cc \textbf{48.7} & \cc \textbf{58.6} \\
    & LaCo$_{24}$             & 44.8 & 84.6 & 44.2 & 25.8 & 29.4 & 61.6 & 51.9 & 43.7 & 65.2 & 23.2 & 41.4 & \textbf{25.1} & 41.2 \\
    & Streamline$_{24}$       & 44.8 & 31.6 & 36.4 & 24.1 & 24.7 & 48.8 & 53.7 & 37.0 & 61.1 & 21.9 & 4.1  & 22.9 & 33.5 \\
    & \cc Ours$_{24}$         & \cc 44.8 & \cc 23.4 & \cc \textbf{54.0} & \cc \textbf{30.0} & \cc \textbf{32.0} & \cc \textbf{62.3} & \cc \textbf{58.4} & \cc \textbf{52.2} & \cc \textbf{70.2} & \cc \textbf{30.0} & \cc \textbf{45.0} & \cc 22.9 & \cc \textbf{45.7}  \\
    \bottomrule
  \end{tabular}
\caption{Experimental results of pruning methods on Qwen models.}
\label{tab:qwen-performance-table}
\end{table*}

\subsubsection{Pruning}
\label{sec:pruning-results}
\cref{tab:llama-performance-table,tab:qwen-performance-table} show the experimental results of pruning methods on LLaMA-family and Qwen models.
Under the same training duration, our method performs favorably compared to other pruning methods.
Notably, our method performs better at higher pruning ratio in most cases, which indicates its stability in model compression.
Moreover, this comparison is particularly meaningful because our method uses training resources comparable to those of current pruning methods.
The results suggest that a well-designed curriculum learning strategy can serve as a unified compression framework for both KD and pruning.
To further validate training stability, we provide the experimental results of standard deviation values in \cref{sec:train-stability}.

\begin{table}[t!]
  \centering
  \scriptsize
  \begin{tabular}{l|c|cccc}
    \toprule
                                          & Train-hour  & OWT & WKT & \multicolumn{2}{c}{LAMBADA} \\
                                          & (H) & ppl & ppl & ppl & acc \\
    \midrule
    GPT-2$_{768 \times 12}$                & >1000 & 23.1             & 37.5            & 35.1         & 46.0   \\
    \midrule
    LWD$_{768 \times 6}$                   & 400 & 29.7             & 51.9            & 91.9         & 22.0   \\
    TED$_{768 \times 6}$                   & 450 & 28.5             & 48.1            & 87.2         & 23.0   \\
    \midrule
    Streamline$_{768 \times 6}$          & 160  & 29.7             & 62.6            & 127.9        & 21.1   \\
    LaCo$_{768 \times 6}$                & 160  & 29.2             & 61.7            & 136.7        & 21.8    \\
    \midrule
    \ourrow Ours$_{768 \times 6}$    & 160 & \textbf{26.4} & \textbf{46.3} & \textbf{69.4}& \textbf{27.9}   \\
    \bottomrule
  \end{tabular}     
\caption{Experimental results regarding the compression methods on GPT-2. OWT and WKT represent OpenWebText and WikiText-103 datasets, respectively.}
\label{tab:gpt2-prune-performance}
\end{table}

\subsubsection{Comparison of KD and pruning}
\cref{tab:gpt2-prune-performance} indicates a comparison of KD and pruning methods.
To ensure a fair comparison, we allocate the same training budget used by our method to the recovery stage of the pruning baselines.
The results present that extending the recovery stage improves the performance of pruning methods, such as LaCo and Streamline, but they still do not match our method's performance.
These findings reinforce our hypothesis that pruning enables rapid model compression; however, its performance does not match that of KD due to the absence of the knowledge transfer process.
Consequently, the proposed layer-wise curriculum learning approach outperforms the pruning baselines under the same training hour budget and also surpasses the KD baselines despite using substantially fewer training resources.

\section{Ablation Study}

\subsection{Feature Caching}
\label{sec:ablation-curriculum-strategy}
\cref{tab:compare-curriculum-strategy} demonstrates that both the proposed curriculum learning schedule and feature caching method improve the overall performance considerably compared to their respective baselines.
The individual application of each component yields notable performance gains over the baseline, but their integration leads to the largest improvement.
This phenomenon is rooted in their complementary roles: the curriculum learning schedule stabilizes the optimization of the cumulative error in the deeper layers, while the feature caching method efficiently resolves the feature misalignment across layers.
Consequently, we combine both of them to facilitate more robust representation quality by effectively mitigating the error accumulation inherent in the independent layer-wise learning approach.

\begin{table}[t!]
  \centering
  \setlength{\tabcolsep}{3pt}
  \scriptsize
  \begin{tabular}{c|cc|ccccc|c}
    \toprule
                               & FC & Sched. & Exam. & Unde. & Reas. & Lang. & Know. & Avg.\\
    \midrule
    \multirow{4}{*}{\shortstack{LLaMA3-3B\\$28 \rightarrow 19$}} & \ding{55} & \ding{55} & 38.5 & 64.3 & 51.4 & 40.9 & 28.2 & 45.4 \\
                               & \ding{55} & \ding{51} & 39.6 & 62.1 & 52.1 & 49.1 & 29.9 & 46.8 \\
                               & \ding{51} & \ding{55} & 40.0 & 63.1 & 52.7 & 55.7 & 29.1 & 47.8 \\
                               & \ding{51} & \ding{51} & 42.1 & 63.4 & 53.5 & 61.4 & 31.0 & 49.6 \\
    \midrule
    \multirow{4}{*}{\shortstack{LLaMA3-8B\\$32 \rightarrow 23$}}  & \ding{55} & \ding{55} & 46.9 & 69.1 & 58.3 & 63.9 & 61.7 & 56.9 \\
                                & \ding{55} & \ding{51} & 47.4 & 70.8 & 58.3 & 67.6 & 60.8 & 57.5 \\
                                & \ding{51} & \ding{55} & 47.9 & 69.2 & 59.3 & 66.8 & 62.0 & 57.9 \\
                                & \ding{51} & \ding{51} & 49.2 & 71.9 & 60.8 & 71.2 & 61.3 & 59.5 \\
    \bottomrule 
  \end{tabular}
\caption{
Ablation study of the components in our proposed method.
We compare two components of the proposed method, such as our curriculum learning schedule (Sched.) and feature caching (FC) method.
}
\label{tab:compare-curriculum-strategy}
\end{table}

\begin{table}[t!]
  \centering
  \setlength{\tabcolsep}{3.5pt}
  \scriptsize
  \begin{tabular}{c|l|ccccc|c}
    \toprule
                               &  Loss func.  & Exam. & Unde. & Reas. & Lang. & Know. & Avg.\\
    \midrule
    \multirow{4}{*}{\shortstack{LLaMA3-8B\\$32 \rightarrow 23$}}  & KLDiv. & 41.4 & 72.3 & 55.4 & 56.8 & 61.2 & 53.6 \\
                                & MSE      & 47.5 & 75.1 & 58.9 & 69.6 & 61.7 & 58.5 \\
                                & Norm.MSE & 48.4 & 72.7 & 59.6 & 69.8 & 61.5 & 58.8 \\
                                & Cosine   & 49.2 & 71.9 & 60.8 & 71.2 & 61.3 & 59.5 \\
    \bottomrule 
  \end{tabular}
\caption{
Experimental results regarding the loss function on the proposed method.
}
\label{tab:compare-loss-func}
\end{table}

\subsection{Loss Function}
\label{sec:ablation-loss-function}
\begin{figure}[t!]
    \subfloat[Cosine similarity loss]{\includegraphics[width=0.5\columnwidth]{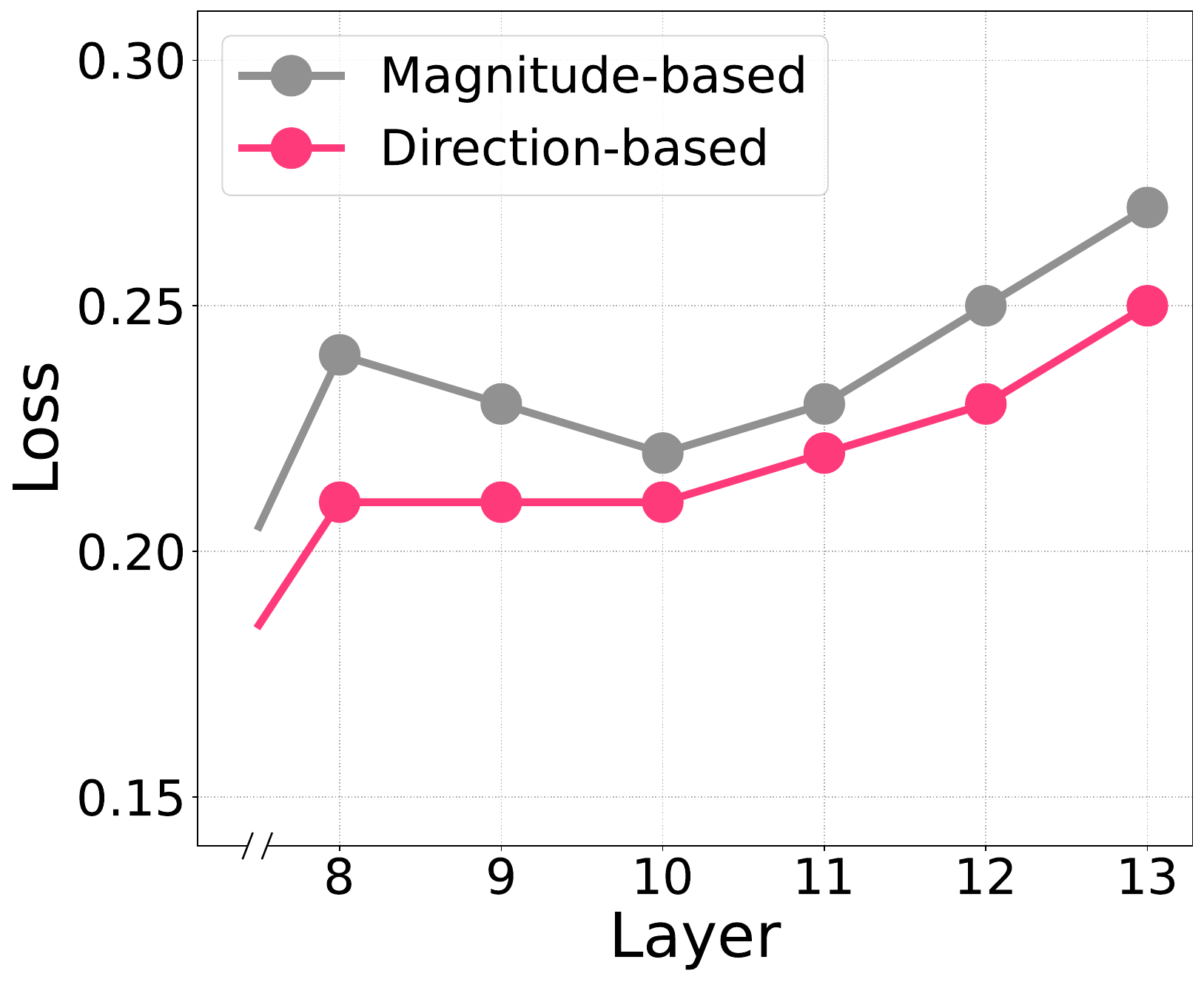}}
    \subfloat[MSE loss]{\includegraphics[width=0.5\columnwidth]{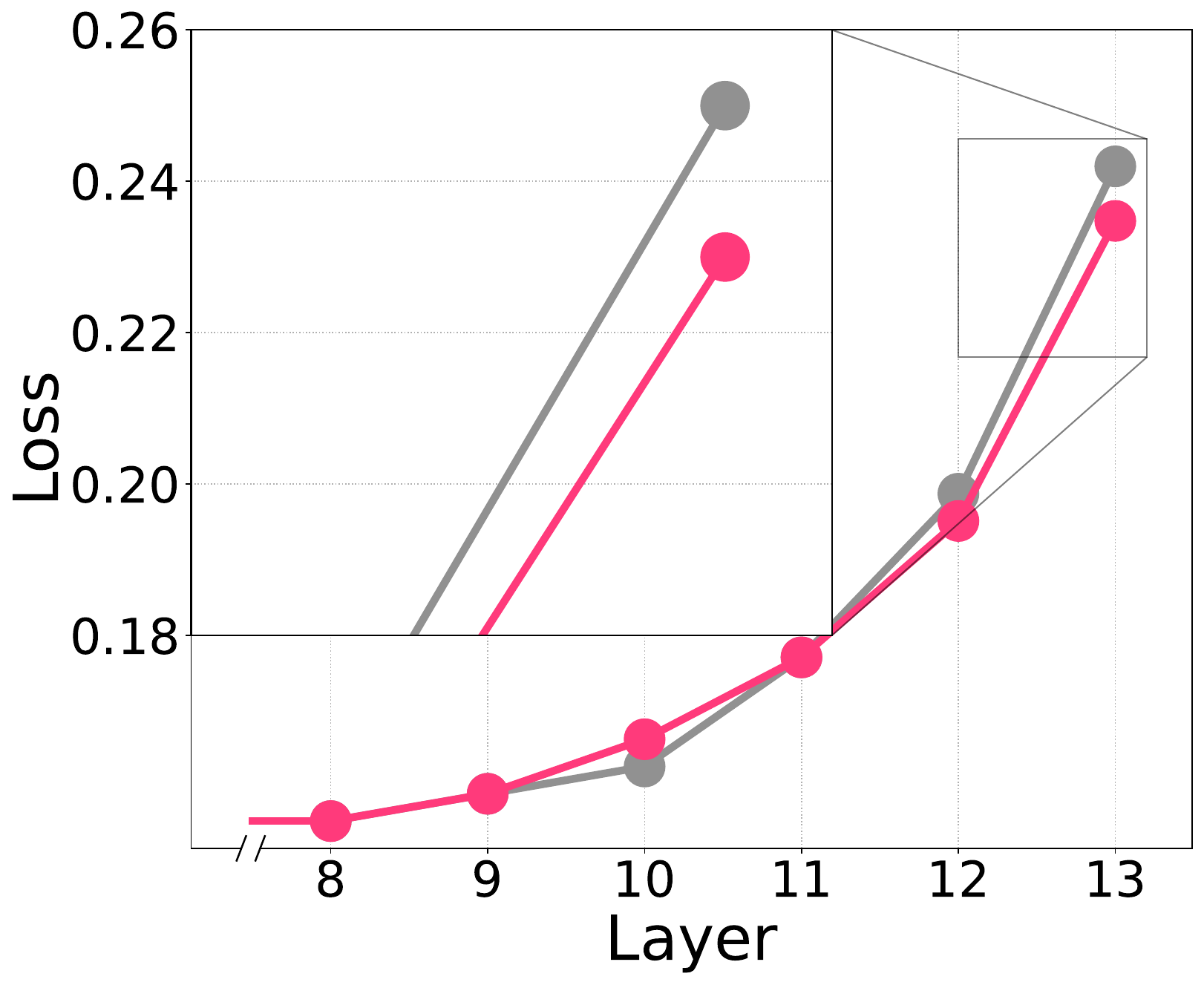}}
    \caption{
    Experimental results of direction-based and magnitude-based objectives.
    We measure (a) cosine similarity loss and (b) MSE loss between teacher and student representations using each objective, respectively.
    }
    \label{fig:compare-loss-function}
\end{figure}

\cref{tab:compare-loss-func} compares direction-based and magnitude-based loss functions.
Direction-based objectives, such as cosine similarity and normalized MSE (Norm.MSE), outperform magnitude-based objectives, including the KL divergence loss (KLDiv.) and MSE.
To further investigate this gap, \cref{fig:compare-loss-function} reports the layer-wise cosine similarity and MSE loss between student and teacher models.
It confirms that direction-based objectives yield better representations under both metrics.
Notably, they achieve lower MSE loss in deeper layers, even though MSE is a magnitude-based metric.
These findings demonstrate that direction-based loss functions are more effective because they explicitly enforce directional alignment, while the normalization layer implicitly stabilizes magnitudes.

\subsection{Recovery Stage}
\cref{tab:compare-recovery} shows the experimental results regarding the recovery stage.
The results demonstrate that the proposed method outperforms the existing SOTA baseline even without any recovery stage, while an extremely short recovery stage yields further performance enhancements.
These findings indicate that the recovery stage serves as a lightweight yet effective training process, whereas the primary performance gains are already established by the proposed curriculum learning approach.
Experimental results of much longer recovery duration are provided in \cref{sec:longer-recovery}.

\begin{table}
  \centering
  \setlength{\tabcolsep}{1.7pt}
  \scriptsize
  \begin{tabular}{c|c|c|ccccc|c}
    \toprule
                            & \makecell{Recov. \\ stage} & \makecell{Train \\ hour} & Exam. & Unde. & Reas. & Lang. & Know. & Avg.\\
    \midrule
    Streamline$_{23}$   & - & 36 & 44.6 & 67.5 & 56.2 & 65.2 & 29.9 & 52.1 \\
    \midrule
    \multirow{2}{*}{\shortstack{Ours$_{23}$}}     & \ding{55} & 32 & 44.9 & 69.2 & 56.5 & 69.2 & 34.7 & 53.4 \\
                                       & \ding{51} & 36  & 45.3 & 76.4 & 58.2 & 71.9 & 31.6 & 54.9 \\
    \midrule
    Streamline$_{16}$    & - & 52  & 30.3 & 61.3 & 41.1 & 7.7 & 23.1 & 34.7 \\
    \midrule
    \multirow{2}{*}{\shortstack{Ours$_{16}$}}   & \ding{55} & 48 & 35.5 & 61.2 & 46.9 & 32.0 & 24.1 & 41.1 \\
                                       & \ding{51} & 52  & 36.8 & 61.6 & 48.8 & 41.7 & 25.5 & 43.4 \\
    \bottomrule 
  \end{tabular}
    \caption{Experimental results regarding recovery stage.}
\label{tab:compare-recovery}
\end{table}

\section{Conclusion}
\label{conclusion}
This paper introduces a novel knowledge transfer method that utilizes a layer-wise curriculum schedule that progressively shifts optimization from easier to harder tasks.
We provide a bound-based analysis of the cumulative error phenomenon inherent in the conventional layer-wise learning approach, which optimizes all layers independently.
Based on the theoretical and empirical evidence, we demonstrate that our curriculum learning approach can effectively mitigate the cumulative error of the deeper layers.
To address feature misalignment across layers efficiently, we develop a feature caching method with a multi-threading strategy that minimizes redundant computation and maximizes resource utilization.
Despite its rapid convergence, our method outperforms conventional knowledge distillation and pruning methods across extensive experiments.
Furthermore, we eliminate the need to load entire models onto each GPU, enabling scalable and efficient training even in memory-constrained environments.
Our results suggest that strong compression performance can be achieved without incurring the conventional training cost throughout optimization.
We believe that the proposed method can serve as a unified compression framework across both knowledge distillation and pruning, thereby enhancing the accessibility of large language models.

\section*{Limitations}
While the proposed framework substantially improves the efficiency of layer-wise learning, it still has two limitations.

First, our recovery stage is implemented as an end-to-end optimization stage over the full model.
This may limit the scalability of the current recovery procedure, especially for much larger models where even a short end-to-end training may become a practical bottleneck.
Developing a more scalable recovery stage that preserves the efficiency of the overall framework remains an important direction for future work.

Second, although our empirical studies cover diverse models and include several large language models, our study does not yet evaluate models with more than 30B parameters.
Extending the evaluation to much larger models is an important direction for future work and would further clarify the scalability of our proposed framework.

\section*{Acknowledgments}
This research was supported by the National Research Foundation of Korea (NRF), Electronics and Telecommunications Research Institute(ETRI), and Institute of Information \& Communications
Technology Planning \& Evaluation (IITP), funded by the Korean government [26CS1100, Development of Proprietary Physical AI-based Small-scale Computers and Integrated Soft Suits], and the Korea government(MSIT) (RS-2024-00356486, RS-2026-25617480, and IITP-2026-RS-2023-00255968).

\bibliography{custom}

\clearpage
\newpage
\appendix

\section{Algorithm of Entire Procedure}
\label{sec:algorithm-of-entire-procedure}
We provide the PyTorch-like pseudocode of layer-wise curriculum learning with feature caching using a multi-threading strategy in \cref{alg:curriculum_know_transfer} to facilitate understanding the entire procedure.
The detailed algorithm is described as follows:
\begin{enumerate}
    \item Partitioning:
    Divide the teacher and student models into $K$ segments.
    Each segment is treated as the basic optimization unit of the layer-wise knowledge transfer procedure.
    For the student model, each segment typically includes a single student layer.
    \item Feature caching from teacher: 
    Forward the teacher model only once to extract intermediate features for all segments.
    These cached teacher features are reused during training as fixed label features.
    The curriculum features are set to the teacher cache at the initial step ($t\!\!\!=\!\!\!0$).
    For large models such as LLaMA-family~\cite{touvron2023llama,grattafiori2024llama} and Qwen~\cite{yang2024qwen25}, the cached features are compressed with LLM.int8()~\cite{10.5555/3600270.3602468} quantization to reduce storage overhead.
    \item Multi-threaded knowledge transfer:
    Each active layer $k \ge \psi(t)$ is assigned to a dedicated device.
    Each worker reads the corresponding cached curriculum and label features, computes the cosine loss, and updates the parameters of the layer.
    Since active layers are independent, their knowledge transfer can be executed in parallel with multi-threading.
    \item Progressive freeze:
    At each curriculum transition, freeze the shallowest active student layer.
    This progressively increases the minimum active depth $\psi(t)$, reallocating the optimization budget from already-converged layers to deeper, more error-prone ones.
    \item Efficient cache update:
    Update the curriculum features using the current student model, but recompute only the active layers.
    The cached outputs of the frozen ones are reused, which avoids redundant forward passes and preserves computational efficiency.
    \item Iteration:
    Repeat the 3) multi-threaded knowledge transfer, 4) progressive freezing, and 5) efficient cache updating until all layers have been optimized and frozen.
    \item Short recovery:
    After the layer-wise knowledge transfer, perform a very short end-to-end recovery stage with LoRA~\cite{hu2022lora} adapters on student model.
    This stage refines alignment across layers at low additional cost by updating only a small set of trainable adapter parameters.
\end{enumerate}

\begin{algorithm*}[t!]
\DontPrintSemicolon
\begin{lstlisting}[basicstyle=\ttfamily\footnotesize]
# Assume 1: n_T > n_S (teacher segments contain more layers than student segments)
# K : Number of segments
# x : Input data
from threading import Thread as thr

# Split teacher and student models into K segments
f_T = [Teacher(layers=k*n_T to (k+1)*n_T) for k in range(K)]
f_S = [Student(layers=k*n_S to (k+1)*n_S) for k in range(K)]

# Cache(update) forward-propagated features
def feature_cache(x, segments, cache_h=None, start_k=0):                 
    if cache_h is None:                 # Initialization
        cache_h = [None] * (K+1)        
        cache_h[0] = x
        
    h = cache_h[start_k]
    for k in range(start_k, K):         # Iteration for active segments
        h = segments[k](h)              # Forward propagation
        cache_h[k+1] = h                # Cache propagated features
    return cache_h

# Knowledge transfer of each layer
def know_transfer(k, pair_h):           
    h_curr = pair_h['curr'][k]          # Load cached curriculum/label features
    h_label = pair_h['label'][k+1]       
    h_output = f_S[k](h_curr)           # Forward with curriculum features
    L_kd = cosine_loss(h_label, h_output)   
    L_kd.backward()                     
    optimizer_layers[k].step()          # Update student layer

teacher_h = feature_cache(x, f_T)       # Initialize curriculum/label features
pair_h = {'curr': teacher_h, 'label': teacher_h}
del f_T                                 # Decommission teacher model

for psi_t in range(K):                  # Curriculum Schedule for active layers
    thrds = [thr(target=know_transfer, args=(k, pair_h)) for k in range(psi_t, K)]
    [t.start() for t in thrds]        
    [t.join() for t in thrds]           # Curriculum learning with multi-threading

    # Freeze the shallowest active layer
    for param in f_S[psi_t].parameters():
        param.requires_grad = False

    # Update curriculum features with forward-propagated ones
    pair_h['curr'] = feature_cache(x, f_S, pair_h['curr'], start_k=psi_t)

student = merge_layers(f_S)             # Merge layers into a full student model
student = attach_lora(student)          # Attach LoRA adapters to student model
short_recover(student)                  # Apply short recovery
        
\end{lstlisting}
\caption{PyTorch-like pseudocode of layer-wise curriculum learning with feature caching using a multi-threading strategy. It illustrates the detailed implementation and execution workflow.}
\label{alg:curriculum_know_transfer}
\end{algorithm*}

\begin{table}[tb]
  \centering
  \setlength{\tabcolsep}{3pt}
  \scriptsize
  \begin{tabular}{c|c|ccccc}
    \toprule
                               & Norm. & Layer 14 & Layer 15 & Layer 16 & Layer 17 & Layer 18 \\
    \midrule
    \multirow{2}{*}{$L^{\text{lower}}_k$} & \ding{55} & 1.09 & 1.20 & 1.15 & 1.36 & 2.66 \\
                                   & \ding{51} & 0.79 & 0.85 & 0.96 & 1.01 & 1.01 \\
    \midrule
    \multirow{2}{*}{$L^{\text{sensi}}_k$}  & \ding{55} & 2.23 & 2.13 & 2.08 & 3.35 & 8.64 \\
                                    & \ding{51} & 0.15 & 0.12 & 0.10 & 0.11 & 0.12 \\
    \bottomrule 
  \end{tabular}
\caption{
Empirical analysis of the local Lipschitz continuity assumption.
$L^\text{lower}_k$ denotes an empirical lower bound on the local Lipschitz constant, while $L^\text{sensi}_k$ represents the local sensitivity of each student layer.
}
\label{tab:empirical-analysis-lipschitz}
\end{table}

\section{Empirical Validation of the Local Lipschitz Continuity Assumption}
\label{sec:empirical-analysis-lipschitz}
In \cref{sec:cumulative-error}, our theoretical analysis assumes that each student layer is locally Lipschitz continuous on the feature space induced by output normalization.
As shown in \cref{eq:mismatch_error_minkowski}, this assumption enables us to relate teacher-student misalignment at the input of a layer to that at the output, which in turn yields the recursive error bound derived in \cref{eq:upper_bound_error_unroll}.
A primary concern is whether this bound scales so large with depth that it becomes vacuous.
To address this concern, we empirically quantify two factors that capture (i) a lower bound of the local Lipschitz constant and (ii) a local sensitivity of each layer under a small perturbation.
We first measure the lower bound of the local Lipschitz constant defined as:
\begin{equation}
\label{eq:lower-bound-lipschitz-constant}
    L^\text{lower}_k = \frac{||f_k(T_{k-1}) - f_k(S_{k-1})||}{||T_{k-1} - S_{k-1}||},
\end{equation}
where $f_k(\cdot)$ represents the $k_{th}$ student layer, while $T_{k-1}$ and $S_{k-1}$ denote $(k\!-\!1)_{th}$ features of teacher and student, respectively.
This quantity is the empirical amplification factor effectively derived from the actual teacher-student feature mismatch.
Second, to directly validate the local Lipschitz stability, we estimate a local sensitivity via a small perturbation on the input:
\begin{equation}
    L^\text{sensi}_k \approx \frac{||f_k(S_{k-1}+\epsilon) - f_k(S_{k-1})||}{||\epsilon||},
\end{equation}
where $\epsilon$ is a perturbation vector with $||\epsilon|| = 10^{-3}$.
While $L^\text{lower}_k$ measures the amplification factor effectively caused by teacher-student mismatch, $L^\text{sensi}_k$ quantifies the local stability of the student layer.
It serves as a direct empirical verification of the local Lipschitz continuity assumption.
As shown in \cref{tab:empirical-analysis-lipschitz}, our empirical results indicate that output normalization substantially suppresses the amplification and ensures stability.
In contrast, under unnormalized settings, both $L^\text{lower}_k$ and $L^\text{sensi}_k$ exhibit significant increases with depth, leading to potential error explosion or gradient instability.
Consequently, it demonstrates that the recursive error bound derived in \cref{eq:upper_bound_error_unroll} is unlikely to explode within the practical learning regime we consider, which directly addresses concerns regarding vacuous bounds.
The depth-wise observation of this empirical validation highlights that the output normalization is a critical mechanism for enforcing local smoothness and mitigating excessive amplification of error during layer-wise learning.

\begin{figure}[t!]
  \centering
  \includegraphics[width=.95\columnwidth]{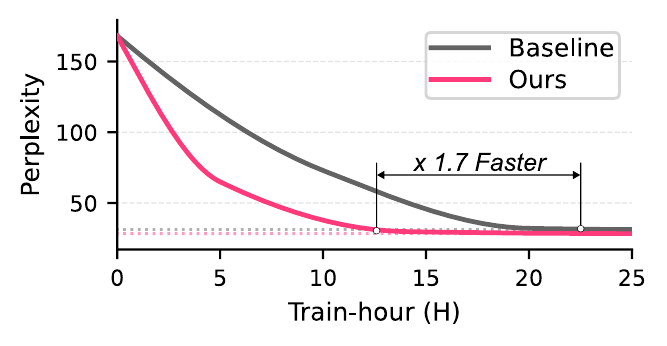}
  \caption{
  Comparative results of the perplexity convergent trend.
  We highlight the advantage of our method in terms of the training hours required to reach a similar perplexity.
  Baseline denotes the knowledge transfer method without our curriculum learning strategy.
  }
  \label{fig:compare-perplexity-trend}
\end{figure}

\section{Detailed Description of \texorpdfstring{$\psi(t)$}{psi(t)}}
\label{sec:psi-description}
The stability of layer-wise curriculum learning schedule is determined through $\psi(t)$, which denotes the index of the shallowest active layer  at timestep $t$.
Accordingly, $k \ge \psi(t)$ layers remain trainable, while layers $k < \psi(t)$ layers are frozen and excluded from subsequent optimization.
Thus, $\psi(t)$ determines which layers remain active and continue to optimize.
This makes $\psi(t)$ a stability control variable rather than a simple indexing schedule.
Furthermore, \cref{eq:upper_bound_layers} makes explicit that the effectiveness of freezing depends on the quality of the frozen layer through the boundary term $e_{\psi(t)-1}$.
In other words, the proposed schedule is effective when the frozen shallow layer has already been sufficiently optimized, so that residual error is small enough for stable optimization.
If shallow layers are frozen too early, $e_{\psi(t)-1}$ remains large and is propagated through deeper layers.
To effectively facilitate our layer-wise curriculum learning, we propose two types of schedules.

\textbf{Static Schedule.}
In our main experiments, we define $\psi(t)$ as a static exponential schedule $\psi(t) = e^{\alpha t} + \beta$, where $\alpha$ controls the rate at which the curriculum progresses over time and $\beta$ determines the overall offset.
It demonstrates that the schedule allocates relatively more timesteps to earlier curriculum stages.
In the implementation, $\psi(t)$ is discretized as an integer value and used to determine the minimum depth of active layers, increasing by one at each curriculum transition.
We set $\alpha = 0.035$ for the schedules of all models.

\textbf{Convergence-aware Adaptive Schedule.}
Although the exponential schedule provides a simple and effective implementation, its curriculum transition times are determined in advance and do not take account of differences in convergence speed across models.
Consequently, we additionally propose a convergence-aware adaptive schedule, in which each curriculum transition is determined directly by the layer-wise optimization state of the shallowest active layer.
With this schedule, the shallowest active layer is frozen only after its layer-wise loss has stabilized, rather than according to a predetermined number of optimization steps.
We regard the shallowest active layer $\psi(t)$ as stabilized once the change in its loss is less than $1e\!-\!4$ over two consecutive epochs.
The adaptive schedule preserves the core optimization principle of our layer-wise curriculum learning approach, while further supporting generalizability across various models.
Ablation studies on schedule types and number of layers frozen per curriculum transition are provided in \cref{sec:ablation-schedule-psi,sec:ablation-num-layers}, respectively.

\begin{table}[t!]
\centering
\setlength{\tabcolsep}{4.5pt}
\scriptsize
\begin{tabular}{l|cccc}
    \toprule
                     & \multicolumn{2}{c}{GPT-2} & \multicolumn{2}{c}{LLaMA3-3B} \\
                     & Mem. (GB) & Train-hour (H) & Mem. (GB) & Train-hour (H) \\
    \midrule
        DDP              & 1.8 & 0.1 & 20.1 &  3.1  \\
        FSDP             & 1.5 & 0.2 & 12.7 &  19.9  \\
        \ourrow Ours     & \textbf{0.3} & 0.1 & \textbf{4.2}  & 3.3  \\
    \bottomrule 
\end{tabular}
\caption{
Experimental results of GPU memory usage and training hours of 5M tokens on DDP, FSDP, and Ours for knowledge transfer. 
The compression ratio is fixed at 60\% and the batch size is set to 1.
}
\label{tab:compare-gpu-parallelism}
\end{table}

\begin{table}[t!]
  \centering
  \setlength{\tabcolsep}{4pt}
  \scriptsize
  \begin{tabular}{c|c|c} 
    \toprule
             & Method & Throughput (token/s) \\ 
    \midrule
    \multirow{3}{*}{LLaMA2-13B} & Dense & 21.3\\ 
                                & LLM-Pruner & 14.4 \\ 
                                & Ours & 28.6 \\
    \midrule
    \multirow{3}{*}{LLaMA3-8B} & Dense & 33.4 \\ 
                               & LLM-Pruner & 20.1 \\ 
                               & Ours & 50.3 \\
    \bottomrule
    \end{tabular}
\caption{Experimental results regarding the throughput of token generation.}
\label{tab:inference-speed}
\end{table}

\section{Computational Efficiency of GPU Utilization}
\label{sec:gpu-utilization}
The key advantage of the proposed method lies in its ability to achieve lower memory usage and faster training speed than conventional distillation methods, while delivering higher accuracy than pruning approaches.
This is made possible by reducing the memory overhead inherent in DDP and eliminating both the sequential execution and communication overhead associated with FSDP when partitioning models across GPUs.
Furthermore, our framework supports both single- and multi-GPU execution without changing the underlying curriculum learning procedure.
In the single-GPU setting, active segments are loaded and optimized sequentially.
When multiple GPUs are available, active segments are distributed across devices and optimized concurrently.
Each segment has an independent optimization problem conditioned on its cached input and target features.
In \cref{tab:compare-gpu-parallelism}, we report memory usage and training hours for knowledge transfer, comparing the conventional knowledge distillation method applying DDP and FSDP with the proposed approach.
In practice, using DDP with knowledge distillation leads to substantial memory consumption, as the full models must be loaded onto each GPU.
Applying FSDP to knowledge distillation, on the other hand, results in a significant increase in communication overhead due to inter-GPU synchronization.
To mitigate this, the teacher model is often fully loaded onto each GPU, rather than distributed, to optimize training hours.
In contrast, the proposed framework avoids such overhead by selectively loading only the required layers of the model to each GPU, resulting in a reduction of approximately 80\% in memory usage.
This distinction explains why the framework can achieve an optimal training resource usage while remaining compatible with both single- and multi-GPU environments.
As a result, this design eliminates the unnecessary memory consumption and significantly accelerates training while improving scalability.
We also provide the inference speed of student models trained by the proposed method.
As shown in \cref{tab:inference-speed}, our method generates more tokens compared to the dense model and LLM-Pruner during the same inference time.
LaCo and Streamline prune layers in the same way as our approach, yielding identical student model structure. 
Consequently, while our framework achieves inference throughput parity with the pruning methods such as LaCo, and Streamline, it significantly outperforms LLM-Pruner, which prunes channels while maintaining the number of layers.

\section{Detailed Experimental Settings}
\label{sec:appendix-experiment-setup}

\subsection{Training Settings}
All experiments are carried out using PyTorch 2.4.0, with FlashAttention-2~\cite{dao2024flashattention} enabled.
For BERT~\cite{devlin2019bert}, We train our method using 120M tokens from Wikipedia\footnote{https://huggingface.co/datasets/wikimedia/wikipedia} and BookCorpus datasets~\cite{zhu2015aligning}.
For GPT-2~\cite{radford2019language}, we train our method using 120M tokens from the OpenWebText~\cite{Gokaslan2019OpenWeb} dataset.
We construct our training data for LLaMA-family and Qwen models from SlimPajama~\cite{shen2023slimpajama} dataset using the data distribution strategy introduced in Sheared-LLaMA~\cite{xia2023sheared}.
For knowledge transfer, we utilize 120M tokens, followed by 60M tokens for the recovery stage.
We train our method on 8 NVIDIA RTX A5000 GPUs, except for LLaMA2-13B and Qwen2.5-14B, which are trained on 2 RTX A6000 GPUs due to larger memory requirements.
\cref{tab:detailed-hyperparameter-setting,tab:detailed-recovery-setting} provide detailed settings of the hyperparameters.
We use AdamW optimizer with ($\beta_1$, $\beta_2$) and a dropout rate of 0.1 in all experiments.

\begin{table}[t!]
  \centering
  \setlength{\tabcolsep}{4.5pt}
  \scriptsize
  \begin{tabular}{c|c|c|c|c|c|c} 
    \toprule
         &  \#B & LR & LR Sche. & Warm-up & W.D. & ($\beta_1$, $\beta_2$) \\ 
    \midrule
    BERT & 256 & 1e-4 & Cosine & 0.05 & 1e-2 & (0.9, 0.999) \\ 
    \midrule
    GPT-2 & 512 & 6e-4 & Linear & 0.05 & 1e-2 & (0.9, 0.98) \\ 
    \midrule
    LLaMA & 1024 & 3e-4 & Linear & 0.005 & 1e-3 &(0.9, 0.95) \\
    \midrule
    Qwen  & 1024 & 3e-4 & Linear & 0.005 & 1e-1 &(0.9, 0.95) \\
    \bottomrule
    \end{tabular}
\caption{Detailed hyperparameter settings for the knowledge transfer. \#B is a batch size and W.D. denotes weight decay.}
\label{tab:detailed-hyperparameter-setting}
\end{table}

\begin{table}[t!]
  \centering
  \setlength{\tabcolsep}{2pt}
  \scriptsize
  \begin{tabular}{c|c|c|c|c|c|c|c} 
    \toprule
         & Epoch & \#B & LR & LR Sche. & Warm-up & W.D. & ($\beta_1$, $\beta_2$) \\ 
    \midrule
    BERT & 3 & 256 & 1e-4 & Cosine & 0.05 & 1e-2 & (0.9, 0.999) \\ 
    \midrule
    GPT-2 & 3 & 512 & 6e-4 & Linear & 0.05 & 1e-2 & (0.9, 0.98) \\ 
    \midrule
    LLaMA2-7B & 1 & 32 & 1e-4 & Linear & 0.005 & 1e-3 & (0.9, 0.95) \\
    \midrule
    LLaMA2-13B & 1 & 32 & 6e-5 & Linear & 0.005 & 1e-3 & (0.9, 0.95) \\
    \midrule
    LLaMA3-3B & 1 & 32 & 1e-4 & Linear & 0.005 & 1e-3 & (0.9, 0.95) \\
    \midrule
    LLaMA3-8B & 1 & 64 & 1e-4 & Linear & 0.005 & 1e-3 & (0.9, 0.95) \\
    \midrule
    Qwen2.5-7B & 1 & 32 & 1e-4 & Linear & 0.005 & 1e-2 & (0.9, 0.95) \\
    \midrule
    Qwen2.5-14B & 1 & 64 & 6e-5 & Linear & 0.005 & 1e-2 & (0.9, 0.95) \\
    \bottomrule
    \end{tabular}
\caption{Detailed hyperparameter settings for the recovery stage.}
\label{tab:detailed-recovery-setting}
\end{table}

\subsection{Evaluation Settings}
We use lm-eval-harness 0.4.3 package~\cite{eval-harness} to evaluate the methods on downstream tasks, except the GLUE~\cite{wang2018glue} benchmark for BERT.
We evaluate our approach against existing KD methods on the GLUE~\cite{wang2018glue} benchmark for BERT, while verifying across the OpenWebText~\cite{Gokaslan2019OpenWeb}, WikiText-103~\cite{merity2016pointer}, and LAMBADA~\cite{paperno2016lambada} datasets for GPT-2.
For the GLUE benchmark, we report Matthews correlation coefficient for COLA~\cite{warstadt2019neural}, F1 score for MRPC~\cite{dolan-brockett-2005-automatically}, Spearman's rank correlation for STSB~\cite{cer2017semeval}, and accuracy for the others.
The performance results of pruning methods are compared across diverse benchmarks, including ARC-E/C~\cite{clark2018think}, RACE~\cite{lai2017race}, BoolQ~\cite{clark2019boolq}, WinoGrande(WnGd)~\cite{sakaguchi2021winogrande}, HellaSwag(HeSW)~\cite{zellers2019hellaswag}, PIQA~\cite{bisk2020piqa}, MathQA~\cite{amini2019mathqa}, COQA~\cite{reddy2019coqa}, and MMLU~\cite{hendrycks2020measuring}.

\subsection{Baseline Provenance}
The performance results for DistilBERT~\cite{sanh2019distilbert}, MiniLM~\cite{wang2020minilm}, and CKABERT~\cite{dasguptaimproving} are taken from the experimental results reported by ~\cite{dasguptaimproving}, while the performance results of DistilGPT~\cite{sanh2019distilbert}, LWD, and TED~\cite{liang2023less} are taken from ~\cite{liang2023less}.
For the pruning comparisons, we reproduce each baseline using its official implementation and configuration.
To minimize the confounding factors from training data, all pruning methods use the same data source, SlimPajama.
Furthermore, we constrain the total training budget to the same range as our method, corresponding to 28-56 GPU $\times$ Hour.
All performance, training hours, and memory measurements are collected under the same GPU environment.
This controlled protocol is intended to isolate differences in the compression algorithms while avoiding discrepancies arising from hardware, data source, or computational budget.

\section{Training Stability}
\label{sec:train-stability}
\cref{tab:standard-deviation-table} presents the standard deviation of experimental results to provide the training stability.
It shows that the standard deviations of our method are comparable to or lower than those of other methods, demonstrating that our method exhibits adequate stability in the model compression process.

\begin{table}
  \centering
  \setlength{\tabcolsep}{6pt}
  \scriptsize
  \begin{tabular}{c|c|cc} 
    \toprule
              & Method & AVG (ratio 25\%) & AVG (ratio 50\%) \\ 
    \midrule
    \multirow{4}{*}{LLaMA3-3B} & LLM-Pruner  & 42.5$_{\pm0.5}$ & 37.2$_{\pm0.4}$ \\ 
                               & LaCo        & 47.0$_{\pm0.6}$ & 39.0$_{\pm0.7}$ \\ 
                               & Streamline  & 49.0$_{\pm0.5}$ & 36.8$_{\pm0.4}$ \\ 
                               & Ours        & 51.5$_{\pm0.1}$ & 40.7$_{\pm0.1}$ \\
    \midrule
    \multirow{4}{*}{LLaMA3-8B} & LLM-Pruner  & 48.4$_{\pm0.1}$ & 40.6$_{\pm0.2}$ \\ 
                               & LaCo        & 55.5$_{\pm0.2}$ & 43.5$_{\pm0.2}$ \\ 
                               & Streamline  & 59.2$_{\pm0.1}$ & 38.5$_{\pm0.4}$ \\ 
                               & Ours        & 62.3$_{\pm0.2}$ & 45.8$_{\pm0.2}$ \\
    \bottomrule
    \end{tabular}
\caption{Experimental results of standard deviation across pruning methods, validating the stability and reliability of our method.}
\label{tab:standard-deviation-table}
\end{table}

\begin{table}[t!]
  \centering
  \setlength{\tabcolsep}{2pt}
  \scriptsize
  \begin{tabular}{c|c|c|ccccc|c}
    \toprule
                               & Method & VRAM & Exam. & Unde. & Reas. & Lang. & Know. & Avg.\\
    
    \midrule
    \multirow{3}{*}{LLaMA3-3B} & POCL  & 14.3 & 38.1 & 63.0 & 49.8 & 57.4 & 25.6 & 46.0 \\
                               & FDD   & 16.5 & 38.8 & 62.2 & 52.2 & 51.7 & 29.3 & 46.8 \\
                               & Ours  & 6.7  & 43.7 & 64.7 & 55.2 & 66.7 & 31.8 & 51.5 \\
    \bottomrule 
  \end{tabular}
\caption{
Comparative results of VRAM (GB) and performance with KD baselines for LLM.
All methods are evaluated under same train data and training hours.
}
\label{tab:comparison-kd-llm}
\end{table}

\section{KD baselines for LLMs}
\cref{tab:comparison-kd-llm} compares our method against LLM distillation approaches to separate the contribution of our curriculum from other forms of distillation.
POCL~\cite{liu2025being} is a curriculum KD method for LLM that schedules training samples by difficulty and progressively introduces harder subsets during distillation, while our approach constructs its curriculum schedule based on optimization difficulty across layers.
FDD~\cite{gong2025beyond} provides a strong layer-wise KD formulation that aligns both intermediate feature trajectories and their first-order changes across layers.
However, its final output distribution matching requires the full teacher and student models to be loaded on each device, incurring substantially higher memory usage in large model settings.
In contrast, our method preserves the efficiency of localized layer-wise optimization while recovering the benefit of knowledge transfer during curriculum learning.
Both KD baselines use LoRA with rank $16$, following the official training configuration of FDD.
The empirical results confirm that our framework not only effectively mitigates the cumulative error, but also enables efficient knowledge transfer under a limited training budget.

\section{Additional Ablation Studies}

\subsection{Types of Curriculum Schedule}
\label{sec:ablation-schedule-psi}
\cref{tab:ablation-schedule-psi} compares the experimental results regarding types of curriculum schedule.
For static schedules, we consider three regimes: a logarithmic schedule, $\psi(t)=\alpha \log (t) + \beta$, which allocates relatively more timesteps to later curriculum stages; a linear schedule, $\psi(t) = \alpha t + \beta$, which allocates timesteps uniformly across all stages; an exponential schedule, $\psi(t) = e^{\alpha t} + \beta$, which allocates relatively more timesteps to earlier curriculum stages. 
The results demonstrate that the exponential schedule outperform the other schedules.
They indicate that when the shallow layers receive insufficient optimization, they remain under-optimized, amplifying error in deeper layers and leading to degradation of final representation quality.

\cref{tab:ablation-static-adaptive} compare the convergence-aware adaptive schedule with the exponential schedule under the same optimization budget.
The empirical results confirm that the adaptive schedule achieves results comparable to those of the exponential schedule.
They also suggest that the core principle of the layer-wise curriculum learning approach generalizes beyond a exponential schedule, as the optimization budget is reallocated toward deeper layers only after shallower layers have stabilized.

\begin{table}[t!]
  \centering
  \setlength{\tabcolsep}{3pt}
  \scriptsize
  \begin{tabular}{c|c|c|ccccc|c}
    \toprule
                                 & Sche. & $\alpha$ & Exam. & Unde. & Reas. & Lang. & Know. & Avg.\\
    
    \midrule
    \multirow{4}{*}{Ours$_{19}$} & Log.   & 1.038 & 43.1 & 63.8 & 54.4 & 63.9 & 31.6 & 50.6 \\
                                 & Linear & 0.067 & 43.3 & 65.2 & 55.0 & 67.1 & 30.9 & 51.3 \\
                                 & Expo.  & 0.038 & 43.6 & 64.0 & 55.3 & 66.6 & 31.7 & 51.4 \\
                                 & Expo.  & 0.035 & 43.7 & 64.7 & 55.2 & 66.7 & 31.8 & 51.5 \\
    \bottomrule 
  \end{tabular}
\caption{
Experimental results regarding the types of schedules (Sche.) and $\alpha$ under the same optimization budget.
Log. denotes the logarithmic schedule, while Expo. indicates the exponential schedule.
}
\label{tab:ablation-schedule-psi}
\end{table}

\begin{table}[t!]
  \centering
  \setlength{\tabcolsep}{4.5pt}
  \scriptsize
  \begin{tabular}{c|c|ccccc|c}
    \toprule
                                 & Sche. & Exam. & Unde. & Reas. & Lang. & Know. & Avg.\\
    
    \midrule
    \multirow{2}{*}{Ours$_{19}$} & Expo. & 43.7 & 64.7 & 55.2 & 66.7 & 31.8 & 51.5 \\
                                 & Adap. & 43.7 & 64.5 & 55.4 & 66.3 & 32.0 & 51.5 \\
    \midrule
    \multirow{2}{*}{Ours$_{23}$} & Expo. & 50.6 & 81.3 & 62.3 & 79.5 & 61.2 & 62.3 \\
                                 & Adap. & 50.7 & 81.4 & 62.8 & 79.0 & 61.4 & 62.5 \\
    \bottomrule 
  \end{tabular}
\caption{
Experimental results regarding the types of schedule (Sche.) under the same optimization budget.
Expo. and Adap. denote the exponential schedule and the convergence-aware adaptive schedule, respectively.
}
\label{tab:ablation-static-adaptive}
\end{table}

\subsection{Number of Layers Frozen per Curriculum Transition}
\label{sec:ablation-num-layers}
\cref{tab:ablation-num-layers} empirically validates how many student layers should be frozen at each curriculum transition by varying the number of layers frozen per transition from 1 to 3.
The results demonstrate that freezing multiple layers at once degrades compression performance.
We attribute this to an abrupt shift of the optimization focus toward deeper layers, which weakens training stability and reduces the effectiveness of curriculum knowledge transfer.
In contrast, freezing a single layer per transition yields the best performance, suggesting that a fine-grained curriculum learning schedule provides more stable transition from shallow to deep optimization.

\begin{table}[t!]
  \centering
  \setlength{\tabcolsep}{4.5pt}
  \scriptsize
  \begin{tabular}{c|c|ccccc|c}
    \toprule
                               & \# Layers & Exam. & Unde. & Reas. & Lang. & Know. & Avg.\\
    
    \midrule
    \multirow{3}{*}{Ours$_{19}$} & 3 & 43.1 & 64.2 & 54.2 & 63.0 & 31.9 & 50.5 \\
                                 & 2 & 43.2 & 64.4 & 54.7 & 64.8 & 32.3 & 51.0 \\
                                 & 1 & 43.7 & 64.7 & 55.2 & 66.7 & 31.8 & 51.5 \\
    \bottomrule 
  \end{tabular}
\caption{
Experimental results regarding the number of layers (\# Layers) frozen at each curriculum transition.
}
\label{tab:ablation-num-layers}
\end{table}

\begin{table}[t!]
  \centering
  \setlength{\tabcolsep}{3.5pt}
  \scriptsize
  \begin{tabular}{c|c|ccccc|c}
    \toprule
                               & Train-hour (H) & Exam. & Unde. & Reas. & Lang. & Know. & Avg.\\
    
    \midrule
    \multirow{5}{*}{Ours$_{23}$} & 8 & 50.6 & 81.3 & 62.4 & 79.5 & 61.2 & 62.3 \\
                                 & 40 & 50.5 & 81.1 & 62.2 & 79.8 & 61.5 & 62.3 \\
                                 & 80 & 50.5 & 81.4 & 62.4 & 79.9 & 61.8 & 62.4 \\
                                 & 160 & 50.9 & 81.2 & 62.4 & 79.8 & 61.6 & 62.5 \\
                                 & 320 & 51.0 & 81.0 & 62.5 & 80.4 & 61.7 & 62.6 \\
    \bottomrule 
  \end{tabular}
\caption{Experimental results regarding longer training durations in the recovery stage.}
\label{tab:longer-recovery}
\end{table}

\subsection{Longer Recovery Duration}
\label{sec:longer-recovery}
\cref{tab:longer-recovery} reports the effect of extending the recovery stage.
While longer recovery yields slight improvements, the overall gains remain modest.
This suggests that extending the recovery budget alone does not substantially improve performance, while the primary performance gains arise from the proposed knowledge transfer.

\end{document}